\documentclass[]{youtu}

\usepackage{mathpazo}
\usepackage{graphicx}
\usepackage{natbib}
\usepackage[utf8]{inputenc}
\usepackage[T1]{fontenc}
\usepackage{url}
\usepackage{hyperref}
\usepackage{cleveref}
\usepackage{booktabs}
\usepackage{amsfonts}
\usepackage{amsmath}
\usepackage{amssymb}
\usepackage{amsthm}
\usepackage{mathtools}
\usepackage{bm}
\usepackage{nicefrac}
\usepackage{microtype}
\usepackage[table]{xcolor}
\usepackage{algorithm}
\usepackage{algorithmic}
\usepackage{multirow}
\usepackage{subcaption}
\usepackage{array}
\usepackage{tabularx}
\usepackage{pifont}
\usepackage{enumitem}
\usepackage[normalem]{ulem}

\newtheorem{theorem}{Theorem}

\theoremstyle{definition}

\theoremstyle{remark}

\newcommand{\methodname}{\texorpdfstring{\ensuremath{\mathrm{MHPO}}}{MHPO}}
\newcommand{\E}{\mathbb{E}}

\newcommand{\cL}{\mathcal{L}}
\newcommand{\cD}{\mathcal{D}}
\newcommand{\KL}{\mathrm{KL}}

\newcommand{\sech}{\mathrm{sech}}

\newcommand{\sg}[1]{\mathrm{sg}\!\left(#1\right)}

\definecolor{dsAIME24}{RGB}{90,140,245}
\definecolor{dsAIME25}{RGB}{140,120,240}
\definecolor{dsAMC23}{RGB}{245,120,170}
\definecolor{dsHMMT25}{RGB}{95,185,160}
\definecolor{dsMATH500}{RGB}{245,150,110}
\definecolor{dsAVG}{RGB}{90,90,90}
\definecolor{dsMathVision}{RGB}{90,140,245}
\definecolor{dsMathVista}{RGB}{245,120,170}
\definecolor{dsMathVerse}{RGB}{95,185,160}

\definecolor{gainpos}{RGB}{128,0,32}
\definecolor{gainneg}{RGB}{0,128,128}
\newcommand{\gainzero}{\,{\color{gray}{\scriptsize (+0.0)}}}
\newcommand{\gainp}[1]{\,{\color{gainpos}{\scriptsize (+#1)}}}

\definecolor{hlGRPO}{RGB}{0,114,178}
\definecolor{hlSHAPE}{RGB}{213,94,0}
\definecolor{hlPhi}{RGB}{0,128,128}       %
\definecolor{hlOmega}{RGB}{148,50,190}     %

\definecolor{hlReff}{RGB}{0,114,178}       %
\definecolor{hlOp}{RGB}{148,50,190}        %
\definecolor{hlPsi}{RGB}{213,94,0}         %
\definecolor{hlBound}{RGB}{0,158,115}      %

\newcommand{\hlbound}[1]{{\color{hlBound}{#1}}}

\definecolor{colDecisive}{RGB}{220,232,252}   %
\definecolor{colDanger}{RGB}{253,226,226}     %
\definecolor{colCaution}{RGB}{255,248,220}    %
\definecolor{colSafe}{RGB}{220,247,220}       %

\newcolumntype{M}{>{\columncolor{dsMATH500!12}\centering\arraybackslash}X}
\newcolumntype{H}{>{\columncolor{dsHMMT25!12}\centering\arraybackslash}X}
\newcolumntype{C}{>{\columncolor{dsAMC23!12}\centering\arraybackslash}X}
\newcolumntype{P}{>{\columncolor{dsAIME25!12}\centering\arraybackslash}X}
\newcolumntype{A}{>{\columncolor{dsAIME24!12}\centering\arraybackslash}X}
\newcolumntype{G}{>{\columncolor{dsAVG!12}\centering\arraybackslash}X}
\newcolumntype{V}{>{\columncolor{dsMathVision!12}\centering\arraybackslash}X}
\newcolumntype{T}{>{\columncolor{dsMathVista!12}\centering\arraybackslash}X}
\newcolumntype{E}{>{\columncolor{dsMathVerse!12}\centering\arraybackslash}X}
\newcolumntype{Y}{>{\centering\arraybackslash}X} %
{
  
  \footnotetext[2]{Corresponding author: kaihanx@hku.hk}
}

\title{MHPO: Modulated Hazard-aware Policy Optimization for Stable Reinforcement Learning}
\author{Hongjun Wang$^{1}$\qquad Wei Liu$^{2}$\qquad Weibo Gu$^2$\qquad Xing Sun$^2$\qquad Kai Han$^1$$^{,\dagger}$ 
}
\affiliation{\textsuperscript{1}The University of Hong Kong \qquad \textsuperscript{2}Tencent Youtu Lab}

\begin{document}

\fancyhead[L]{%
  \includegraphics[height=2.0\baselineskip]{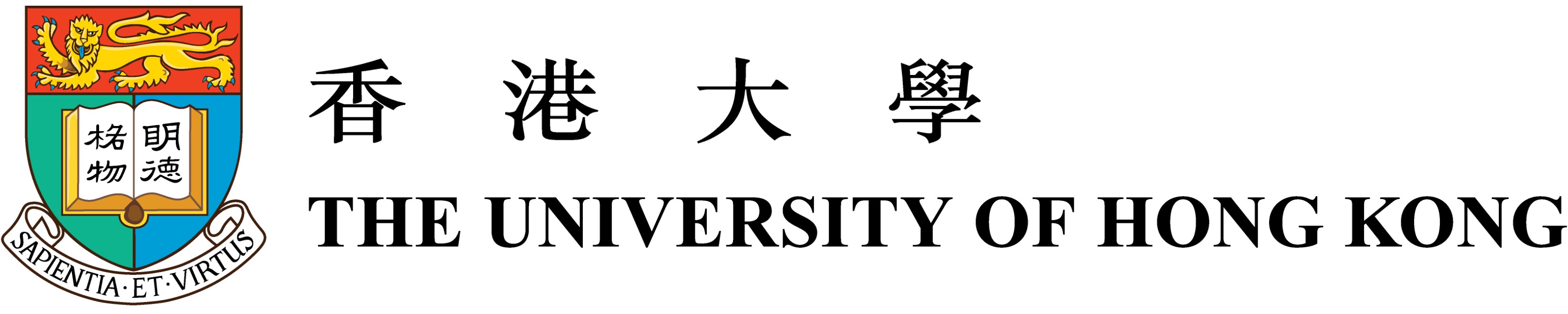}%
}
\fancyhead[R]{\includegraphics[height=2.0\baselineskip]{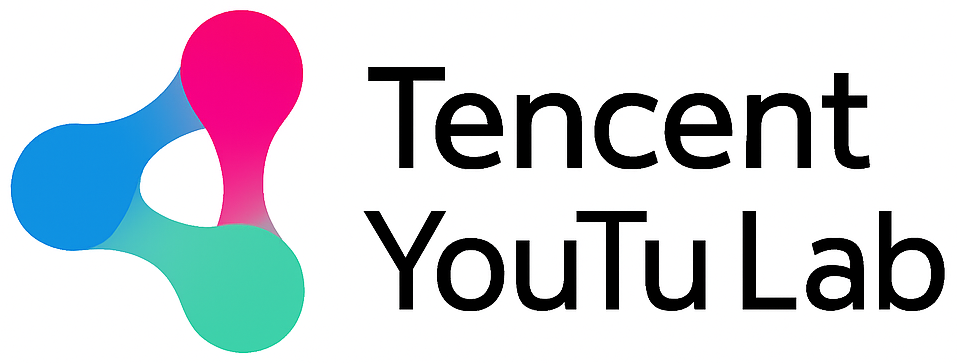}}

\abstract{

Regulating the importance ratio is critical for the training stability of Group Relative Policy Optimization (GRPO) based frameworks. However, prevailing ratio control methods, such as hard clipping, suffer from non-differentiable boundaries and vanishing gradient regions, failing to maintain gradient fidelity. Furthermore, these methods lack a hazard-aware mechanism to adaptively suppress extreme deviations, leaving the optimization process vulnerable to abrupt policy shifts. To address these challenges, we propose \textbf{M}odulated \textbf{H}azard-aware \textbf{P}olicy \textbf{O}ptimization (MHPO), a novel framework designed for robust and stable reinforcement learning. The proposed MHPO introduces a Log-Fidelity Modulator (LFM) to map unbounded importance ratios into a bounded, differentiable domain. This mechanism effectively prevents high-variance outlier tokens from destabilizing the loss landscape while ensuring global gradient stability. Complementarily, a Decoupled Hazard Penalty (DHP) integrates cumulative hazard functions from survival analysis to independently regulate positive and negative policy shifts. By shaping the optimization landscape with hazard-aware penalties, the proposed MHPO achieves fine-grained regulation of asymmetric policy shifts—simultaneously mitigating mode collapse from over-expansion and preventing policy erosion from catastrophic contraction within a stabilized trust region. Extensive evaluations on diverse reasoning benchmarks across both text-based and vision-language tasks demonstrate that MHPO consistently outperforms existing methods, achieving superior performance while significantly enhancing training stability.
}

\maketitle

\begin{figure*}[t]
\centering
\includegraphics[width=\textwidth]{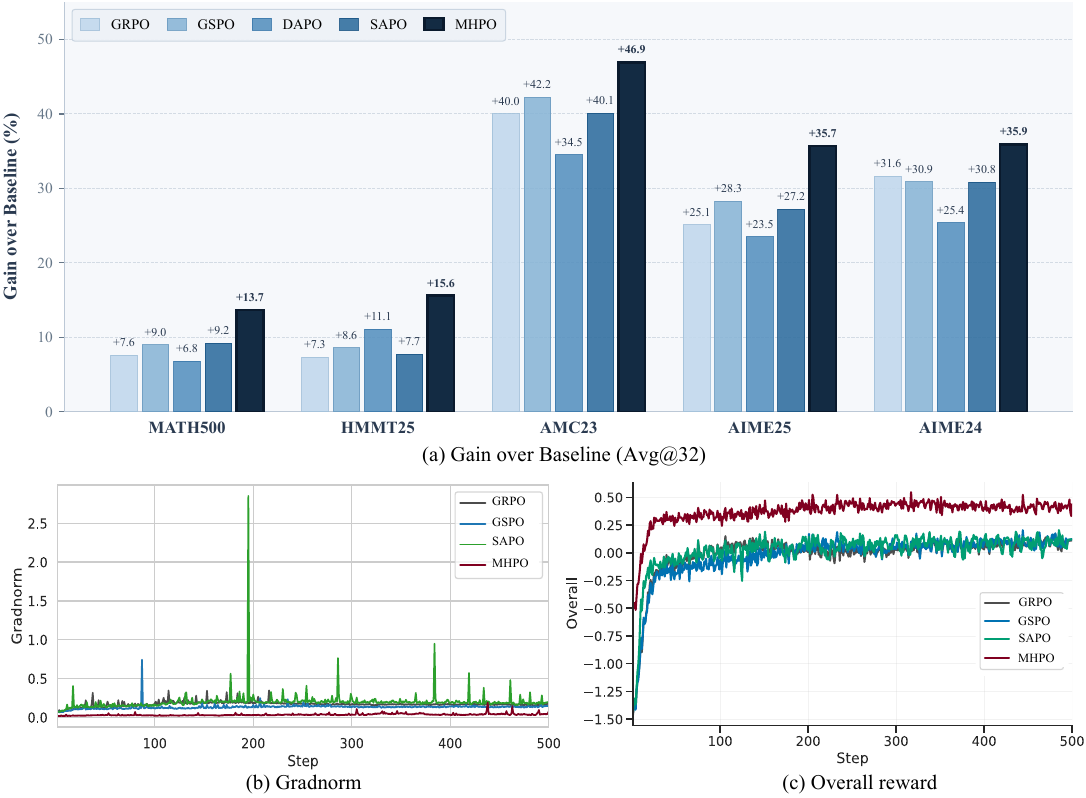}
\caption{Overview of \methodname{}.
\textbf{(a)}~Performance gain over baseline (Avg@32, \%) on Qwen3-4B-Base. The bar chart displays the improvement ($\Delta$) achieved by different RL methods across five benchmarks. \methodname{} (navy) consistently achieves the largest gain on every benchmark.
\textbf{(b)}~Gradient norm trajectory during RL training. Baseline methods exhibit frequent gradient spikes, whereas \methodname{} maintains consistently low and stable gradient magnitudes throughout training, empirically confirming the bounded gradient multiplier guaranteed by our theoretical analysis.
\textbf{(c)}~Overall reward curve during training. \methodname{} attains higher reward earlier and sustains this advantage, while competing methods plateau or exhibit degradation in later stages.}
\label{fig:overview}
\end{figure*}

\section{Introduction}
\label{sec:intro}

Reinforcement learning (RL) has emerged as a pivotal paradigm for the post-training of foundation models, driving significant improvements across both pure text and broader multimodal architectures~\citep{rlhf, o1, guo2025deepseek, llava_rlhf2024, vlm_rlaif2024}. Notably, methods based on Group Relative Policy Optimization (GRPO)~\citep{deepseekmath, guo2025deepseek} have demonstrated that RL can unlock extended chain-of-thought reasoning for complex mathematical and logical problems. Similar performance gains are increasingly pursued for vision-language models~\citep{r1_vlm2025, vision_r12025, vlm_r12025}. 

Despite the remarkable success of GRPO-based methods, stabilizing the training process remains a non-trivial challenge. The importance ratio~\citep{schulman2017proximal}, used to compensate for the discrepancy between the current and reference policies, introduces profound numerical instability. Token-level ratios frequently exhibit extreme variance, an issue significantly exacerbated in long-form Chain-of-Thought (CoT) generation where sequence lengths can extend to thousands of tokens. In such scenarios, the multiplicative accumulation of ratios across extensive sequences can fluctuate across multiple orders of magnitude. These high-variance ``outlier'' tokens trigger massive gradient spikes that destabilize the loss landscape, leading to severe training instability. 

To enhance training stability, most previous works rely on clipping-based methods to constrain the importance ratio within a predefined trust region. For instance, PPO~\citep{schulman2017proximal} and GRPO~\citep{deepseekmath, guo2025deepseek} employ a symmetric hard clip within $[1-\epsilon, 1+\epsilon]$, whereas DAPO~\citep{dapo2025} utilizes more flexible, asymmetric boundaries $[1-\epsilon_{low}, 1+\epsilon_{high}]$. Despite their differences, all such methods inevitably introduce gradient discontinuities and vanishing gradient regions, consequently destabilizing the optimization and preventing tokens outside the trust region from contributing to the learning process. While SAPO~\citep{sapo2025} introduces a soft gating mechanism to maintain gradient smoothness, it fails to decouple the distinct risks associated with directional policy updates. 

Beyond merely constraining the magnitude of the importance ratio, achieving fine-grained control over the asymmetric behavior of directional \emph{policy shifts} (i.e., positive and negative shifts) is fundamentally important for stabilizing policy optimization. A \emph{positive shift} indicates that the current policy increases a token's probability relative to the reference policy, thereby facilitating exploration of new behaviors, whereas a \emph{negative shift} decreases the token probability relative to the reference policy, suppressing undesired behaviors. In the context of policy optimization, the risks associated with probability mass expansion and contraction are inherently asymmetric: overly aggressive positive shifts may induce mode collapse by over-optimizing for a narrow subset of high-reward tokens, while overly aggressive negative shifts---often triggered by high-variance advantage estimates---can catastrophically suppress valid linguistic patterns, causing irreversible policy erosion across subsequent optimization iterations.

To address these limitations, we propose \textbf{M}odulated \textbf{H}azard-aware \textbf{P}olicy \textbf{O}ptimization (\methodname{}), a unified framework designed to guarantee global differentiability while providing fine-grained, hazard-aware control over policy shifts. \methodname{} is composed of two key components: a Log-Fidelity Modulator (LFM) and a Decoupled Hazard Penalty (DHP). Specifically, the LFM employs a scaled $\tanh$ transformation in log-space to map unbounded importance ratios into a bounded, differentiable manifold. This approach ensures a high-fidelity optimization process by preserving the standard policy gradient characteristics near the on-policy anchor while smoothly attenuating the influence of outlier tokens. Complementing the LFM, the DHP introduces a hazard-aware penalty mechanism to independently regulate positive and negative policy shifts. Drawing inspiration from reliability theory and survival analysis, the DHP employs the cumulative hazard function of the Weibull distribution to shape the optimization landscape. This approach facilitates safe exploration by maintaining negligible penalties within a trust region, while triggering rapid penalty acceleration beyond this threshold to suppress large deviations that might destabilize the system. By employing distinct hyperparameter sets, the DHP enables asymmetric hazard shaping, allowing for fine-grained control over the dual regulation of policy shifts. Consequently, with the proposed LFM and DHP, our \methodname{} simultaneously maintains training stability and achieves better performance, as illustrated in Figure.~\ref{fig:overview}. 

In summary, the key contributions of this work are:
\begin{itemize}
\item The Log-Fidelity Modulator (LFM), which employs a scaled $\tanh$ transformation in log-space to map unbounded importance ratios into a bounded, differentiable manifold. By acting as a continuous gradient modulator, the LFM prevents extreme policy shifts from dominating the parameter updates while preserving gradient fidelity.
\item The Decoupled Hazard Penalty (DHP), a novel hazard-aware penalty mechanism inspired by survival analysis and reliability theory. By using cumulative hazard functions, DHP enables asymmetric regulation of positive and negative policy shifts. This allows for fine-grained control over the optimization landscape, encouraging safe exploration while suppressing catastrophic policy erosion.
\item The Modulated Hazard-aware Policy Optimization (MHPO) framework that addresses the inherent instabilities of GRPO-based training. Extensive evaluations across diverse benchmarks, including pure text-based logical reasoning and multimodal vision-language tasks, demonstrate that MHPO consistently outperforms state-of-the-art baselines in both performance and training stability.
\end{itemize}

\section{Related Work}
\label{sec:related}

\subsection{Policy Optimization for Large Language Models}

Reinforcement Learning from Human Feedback (RLHF)~\citep{rlhf} and Proximal Policy Optimization (PPO)~\citep{schulman2017proximal} form the foundation of LLM post-training.
PPO relies on a learned critic that can introduce substantial optimization burden in long chain-of-thought settings~\citep{yuan2025s, kazemnejad2024vineppo}, motivating critic-free alternatives such as Group Relative Policy Optimization (GRPO)~\citep{deepseekmath}, which derives group-based advantages from multiple rollouts per prompt.

A central challenge within the GRPO framework is the instability of importance ratios, and several methods have been proposed to address this issue.
DAPO~\citep{dapo2025} introduces asymmetric clipping boundaries to impose distinct constraints on upward and downward policy shifts.
SAPO~\citep{sapo2025} replaces hard clipping with a sigmoid-based soft gate to restore gradient continuity, though without provable tail attenuation.
GSPO~\citep{gspo2025} elevates ratio control to the sequence level, reducing token-level variance at the cost of fine-grained credit assignment.
However, each of these methods addresses only one facet of the ratio-control problem: clipping enforces boundedness but sacrifices gradient fidelity, soft gating restores smoothness but lacks principled damping guarantees, and sequence-level control suppresses extremes but loses per-token granularity.
A complementary line of work recognizes that positive and negative policy updates exhibit fundamentally different dynamics and require asymmetric treatment.
TOPR~\citep{topr2025} and \citet{asymre2025} stabilize off-policy learning by asymmetrically tapering importance weights across reward polarities.
ASPO~\citep{aspo2025} further observes that importance ratios scale inversely for positive-advantage and negative-advantage tokens and proposes flipping the positive-branch ratio to correct this imbalance.
At the advantage level, A3PO~\citep{a3po2025} introduces adaptive token-level advantage shaping to account for distinct sample polarities, and NGRPO~\citep{ngrpo2025} combines advantage calibration with asymmetric clipping to leverage learning signals from homogeneously incorrect groups.
These methods achieve directional control by modifying the importance weight or the advantage signal, yet none provides smooth, theoretically bounded attenuation directly at the gradient level.

\methodname{} addresses this gap by operating at the gradient multiplier level, unifying fidelity and damping at token granularity without altering the advantage or the importance ratio.
Our approach is also orthogonal to methods that improve other GRPO pipeline components, including \textit{rollout-centric} methods that enhance sample quality or diversity~\citep{noisyrollout2025, visionmatters2025, sharegrpo2025, stepgrpo2025, sophiavlr12025, hintgrpo2025, mgrpo2025, grpocare2025}, \textit{reward-centric} methods that refine verifiable rewards~\citep{visualrft2025, visionreasoner2025, grpo_lead2025}, and \textit{advantage-centric} methods that redesign normalization~\citep{gpg2025, seedgrpo2025, krpo2025}.
These approaches modify the reward or advantage pathway while leaving ratio-induced gradient scaling unchanged, which is the specific focus of \methodname{}.

\subsection{Trust Region Methods in Reinforcement Learning}

The ratio-control methods discussed above can be viewed as implicit trust-region mechanisms that bound policy drift through the importance ratio.
More broadly, trust-region methods constrain policy updates to ensure stable learning~\citep{schulman2015trpo}, and the central challenge lies in balancing constraint satisfaction with computational efficiency.
Two main approaches have emerged in the literature, namely explicit KL-based constraints and implicit ratio-based mechanisms.

KL-based trust regions constrain the policy distribution via a divergence such as $\KL(\pi_{\theta_{\text{old}}}\,\|\,\pi_\theta)$ and are widely used in LLM post-training as both a stabilizer and an implicit prior toward a reference policy~\citep{ziegler2019fine, korbak2022rl, zheng2023secrets}.
However, the effectiveness of KL control depends critically on how KL is approximated and differentiated.
Common estimators exhibit markedly different bias and variance properties, and naive gradient estimation can be brittle~\citep{schulman2020kl, tang2025few, amini2025better}.
Recent analyses further reveal that KL regularization can be unreliable under heavy-tailed misspecification and may induce failure modes when treated as a universal safeguard~\citep{kwa2024catastrophic,vassoyan2025ignore}.
Alternative divergence constraints have also been explored~\citep{wangbeyond}.

Ratio-based mechanisms, exemplified by PPO clipping, approximate the trust region via per-sample ratio constraints rather than explicit divergence computation.
This formulation is practical at scale but introduces gradient discontinuities and binary on/off control, failing to satisfy the fidelity and damping requirements discussed above.
Recent efforts explore smoother surrogates, including bounded log-ratio operators, policy smoothing, and importance-weight truncation, yet these typically address only one of the two requirements in isolation.
\methodname{} targets the log-ratio quantity that locally parameterizes both likelihood change and KL divergence.
The gradient multiplier is smooth and differentiable everywhere while exhibiting principled tail decay that attenuates extreme deviations, thereby subsuming the operational goals of both KL-based and ratio-based trust-region methods within a unified mechanism~\citep{peters2010relative, peters2011reinforcement, zhang2025design}.

\section{Methodology}
\label{sec:method}

In this section, we introduce \methodname{}, a novel reinforcement learning framework designed to enhance training stability in policy optimization. The proposed \methodname{} is composed of a Log-Fidelity Modulator (LFM) and a Decoupled Hazard Penalty (DHP). The LFM maps unbounded importance ratios into a symmetric, differentiable log-space, effectively ensuring global gradient stability and preventing high-variance ``outlier'' tokens from destabilizing the loss landscape. Complementarily, the proposed DHP decouples the regulation of positive and negative policy shifts. This allows for directional penalty shaping that strictly suppresses excessive deviations while maintaining high-fidelity gradient flow within a defined trust region. By integrating these two components, \methodname{} achieves a robust optimization process that simultaneously maintains training stability and achieves better performance. For the rest of this section, we begin by establishing the necessary preliminaries in Section~\ref{sec:prelim_grpo}, followed by a detailed introduction of the \methodname{} objective in Section~\ref{sec:objective}. Finally, Section~\ref{sec:stability} provides a rigorous theoretical analysis to validate the stability of our proposed framework.

\subsection{Preliminaries of GRPO}
\label{sec:prelim_grpo}

Let $\pi_{\theta}$ denote an autoregressive language model parameterized by $\theta$. Given a query $p$ from the query set $\cD$, GRPO~\citep{deepseekmath} samples a group of $K$ responses $\{q^{1}, \ldots, q^{K}\}$ from the reference policy $\pi_{\theta_{\text{old}}}$. The likelihood of each response $q^{i}$ is factorized as $\pi_{\theta_{\text{old}}}(q^{i} \mid p)$ = $\prod_{t=1}^{T_i} \pi_{\theta_{\text{old}}}(q_t^{i} \mid p, q_{<t}^{i})$, where $T_i$ denotes the sequence length. Each response $q^{i}$ is then evaluated by a reward model or a verifiable rule to obtain a scalar reward $R^{i}$. 

Following \citep{deepseekmath}, the GRPO training objective is defined at the token level as:
\begin{equation}
    \cL_{\mathrm{GRPO}}(\theta)
    =
    -\E_{p\sim\cD,\ \{q^{i}\}_{i=1}^{K}\sim\pi_{\theta_{\text{old}}}(\cdot\mid p)}
    \left[
        \frac{1}{K}\sum_{i=1}^{K}\frac{1}{T_i}\sum_{t=1}^{T_i}
        \min\!\Big(r_{t}^i(\theta)\,\hat{A}^{i}_t,\ \mathrm{clip}\!\big(r_{t}^i(\theta),\,1{-}\epsilon,\,1{+}\epsilon\big)\,\hat{A}^{i}_t\Big)
    \right],
    \label{eq:grpo_loss_full}
\end{equation}
where
\begin{equation}
r_t^{i}(\theta) = \frac{\pi_\theta(q_t^i \mid p, q_{<t}^i)}{\pi_{\theta_{\text{old}}}(q_t^i \mid p, q_{<t}^i)},~~\hat{A}^{i}_t = \hat{A}^{i} =  \frac{R^{(i)} - mean(\{R^{1}, \ldots, R^{K}\})}{std(\{R^{1}, \ldots, R^{K}\})},~~\epsilon > 0.
\end{equation}
Here $\epsilon$ is the clipping threshold, $r_t^{i}(\theta)$ is the importance ratio measuring how much the current policy has deviated from the reference policy at token position t, and $\hat{A}^{i}_t$ is the group-relative advantage sharing within the group of $K$ responses.

\paragraph{Positive and Negative Policy Shifts.}
The importance ratio $r_t^i(\theta)$ characterizes the token-level \emph{direction} and \emph{magnitude} of the policy shift relative to the reference policy $\pi_{\theta_{\text{old}}}$. Specifically, a \emph{positive shift} ($r_t^i(\theta)>1$, equivalently $\log r_t^i(\theta)>0$) indicates that the current policy assigns higher probability to token $q_t^i$ than the reference policy, whereas a \emph{negative shift} ($r_t^i(\theta)<1$) indicates a probability decrease. This shift direction is independent of the advantage sign $\hat{A}^{i}_t$, which determines whether the gradient ultimately reinforces ($\hat{A}^{i}_t>0$) or suppresses ($\hat{A}^{i}_t<0$) the sampled token; the two signals jointly govern the final update dynamics through their product in Eq.~\eqref{eq:grpo_loss_full}. Critically, the two shift directions have asymmetric risk profiles: overly large positive shifts can over-amplify a narrow subset of tokens/trajectories and induce mode collapse, while overly large negative shifts can catastrophically suppress broad linguistic patterns and cause irreversible policy erosion or collapse. This motivates decoupled, direction-aware regularization (Section~\ref{sec:DHP}) that can apply distinct controls to the $r>1$ and $r<1$ regimes.

\subsection{The \methodname{} Objective}
\label{sec:objective}

We propose the \methodname{} objective, which is designed to enhance training stability and provide fine-grained control over gradient updates. The model parameters $\theta$ are optimized by minimizing the following objective function:
\begin{equation}
    \cL_{\methodname{}}(\theta) =
    -\E_{p\sim\cD,\ \{q^{i}\}_{i=1}^{K}\sim\pi_{\theta_{\text{old}}}(\cdot\mid p)}
    \Bigg[
        \frac{1}{K}\sum_{i=1}^{K}\frac{1}{T_i}\sum_{t=1}^{T_i}
        \exp\Big(\psi(r_{t}^i(\theta)) - \zeta(r_{t}^i(\theta))\Big) \hat{A}^{i}_t
    \Bigg],\label{equation:methodname_loss}
\end{equation}
where the transformation functions $\psi(\cdot)$ and $\zeta(\cdot)$ are defined as:
\begin{equation}
    \psi(\cdot) = c\tanh\frac{\log (\cdot)}{c},~~\zeta(\cdot) = \Big(\frac{s(\psi(\cdot))}{\lambda_{+}}\Big)^{k_+} + \Big(\frac{s(-\psi(\cdot))}{\lambda_{-}}\Big)^{k_-}.
    \label{equation:psi_zeta_definition}
\end{equation}
Here, $s(\cdot)$ denotes the standard softplus function, while $c$ and $(k_+, k_-, \lambda_+, \lambda_-)$ represent the predefined hyper-parameters.
We next elaborate on the two principal components of this objective, namely the Log-Fidelity Modulator and the Decoupled Hazard Penalty.

\subsubsection{The Log-Fidelity Modulator}

To address the instability inherent in raw importance ratio optimization, we introduce the Log-Fidelity Modulator (LFM). The LFM operator
\begin{equation}
\psi(\cdot) = c \tanh\!\left(\frac{\log (\cdot)}{c}\right)
\end{equation}
first projects the importance ratio $r_{t}^i(\theta)$ into a symmetric log-space to ensure training stability and balanced sensitivity. By operating in the logarithmic domain rather than the raw ratio space, the LFM transforms the multiplicative nature of policy updates into a linear domain where $\log(r_{t}^i(\theta)) = 0$ (i.e., $r_{t}^i(\theta) = 1$) serves as the on-policy anchor. This formulation allows the model to treat positive ($r_{t}^i(\theta) > 1$) and negative ($r_{t}^i(\theta) < 1$) policy updates with mathematical parity while enhancing numerical stability against high-variance gradients.

In contrast to the non-differentiable ``hard'' clipping used in standard GRPO—which often suffers from abrupt gradient vanishing and discontinuities at the trust region boundaries—the LFM serves as a smooth saturation layer via a scaled $\tanh$ function, as illustrated in Figure.~\ref{fig:eim_analysis}(a). Specifically, as the importance ratio deviates significantly from the anchor and the log-ratio approaches $\pm\infty$, the modulator effectively constrains the contribution of any individual sample to the objective function within the closed interval $[-c, c]$. From an optimization perspective, this transformation serves as a continuous gradient modulator: the derivative of the LFM  (see Figure.\ref{fig:eim_analysis}(b)) systematically attenuates the influence of tokens with extreme importance ratios while preserving a non-vanishing gradient signal. By preventing massive outliers from dominating the parameter updates without sacrificing differentiability, the LFM ensures a consistent gradient flow across the entire domain, thereby enhancing the numerical stability and robustness of the convergence process.

To analyze the optimization dynamics, we examine the gradient behavior of the LFM operator:
\begin{equation}
    \frac{d\psi}{dr_t^i(\theta)} = \frac{d}{dr_t^i(\theta)} \left[ c \tanh\left(\frac{\log r_t^i(\theta)}{c}\right) \right] = \frac{1}{r_t^i(\theta)} \text{sech}^2\left(\frac{\log r_t^i(\theta)}{c}\right).
\end{equation}
Based on this derivative, the LFM operator exhibits three distinct properties that facilitate stable policy optimization:

\begin{figure}[t]
\centering
\includegraphics[width=\linewidth]{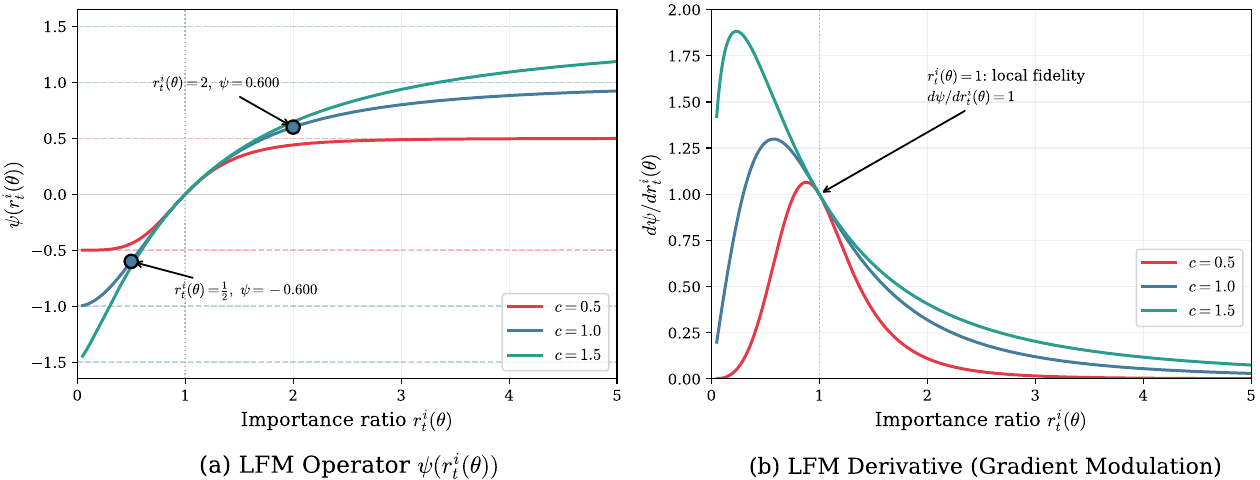}
\caption{Characteristics of the Log-Fidelity Modulator (LFM). \textbf{(a)}~The LFM operator $\psi(r_t^i(\theta))$ with different values of $c$. The transformation exhibits reciprocal antisymmetry ($\psi(1/r_t^i(\theta)) = -\psi(r_t^i(\theta))$) around the on-policy anchor ($r_t^i(\theta)=1$) and smoothly bounds the output within the interval $[-c, c]$. The marked points at $r_t^i(\theta) = \tfrac{1}{2}$ and $r_t^i(\theta) = 2$ (for $c=1.0$) confirm $\psi(\tfrac{1}{2}) = -\psi(2)$. 
\textbf{(b)}~The derivative of the LFM operator, demonstrating smooth gradient attenuation for extreme importance ratios while preserving local fidelity ($d\psi/dr = 1$) near the on-policy anchor.}
\label{fig:eim_analysis}
\end{figure}

\begin{itemize}
    \item \textbf{High-Fidelity Local Mapping} (P1). In the regime where the importance ratio is near the on-policy anchor ($r \approx 1$, or $|\log r| \ll c$), the LFM operator preserves the characteristics of standard policy gradients. Using the first-order Taylor expansion $\tanh(x) \approx x$ for $x \to 0$, the LFM operator acts as an identity mapping in the logarithmic domain:
    \begin{equation}
    \psi(r_t^i(\theta)) \approx c \cdot \frac{\log r_t^i(\theta)}{c} = \log r_t^i(\theta).
    \end{equation}
    Consequently, for small policy updates, the derivative $d\psi/dr_t^i(\theta) \approx 1/r_t^i(\theta)$ recovers the standard log-likelihood formulation. During stable training phases or the early stages of optimization, the LFM operator maintains high fidelity to the original gradient magnitude and direction (indicated by $d\psi/dr_t^i(\theta) = 1$ at $r_t^i(\theta)=1$ in Figure.~\ref{fig:eim_analysis}(b)), guaranteeing efficient learning without introducing bias for on-policy samples.
    \item \textbf{Smooth Gradient Attenuation} (P2). As the policy deviates significantly from the reference, the LFM operator provides global regularization by modulating the first-order gradient magnitude. As $r_t^i(\theta) \to \infty$ or $r_t^i(\theta) \to 0$, the exponential decay of the $\text{sech}^2(\cdot)$ term dominates the expression, forcing the gradient to vanish gracefully. Crucially, unlike hard-clipping mechanisms that truncate updates entirely, the LFM operator ensures that extreme samples still contribute to the gradient descent process without overpowering it, thereby preventing high-variance ``outlier'' tokens from destabilizing the global loss landscape. 

    \item \textbf{Higher-Order Differentiability} (P3). Beyond gradient magnitude, the LFM operator offers superior optimization characteristics through its higher-order differentiability ($C^\infty$ regularity). While standard GRPO suffers from gradient discontinuities—where derivatives drop abruptly to zero at boundaries—these ``mathematical shocks'' often destabilize the momentum buffers in adaptive optimizers (such as Adam~\citep{kingma2015adam}). Through the combined effects of P2 and P3, the first and second-order moments in adaptive optimizers remain stable and consistent. This facilitates a more robust and predictable convergence process even in the presence of extreme importance ratios.
\end{itemize}

\subsubsection{The Decoupled Hazard Penalty}
\label{sec:DHP}

\begin{figure}[t]
\centering
\includegraphics[width=\linewidth]{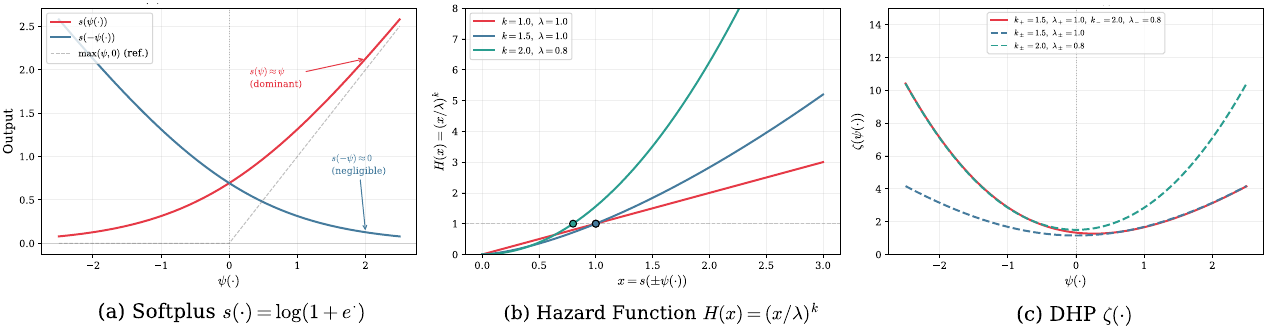}
\caption{Analysis of the Decoupled Hazard Penalty (DHP). \textbf{(a)}~The penalty $\zeta(\psi)$ with different $(\lambda, k)$ configurations. The default asymmetric setting ($\lambda_+=1.0, k_+=1.5, \lambda_-=0.8, k_-=2.0$) applies stronger suppression for negative shifts. \textbf{(b)}~The corresponding survival weight $w = \exp(-\zeta) \in (0, 1]$, demonstrating that the DHP can only attenuate and never amplify token contributions.}
\label{fig:dhp_analysis}
\end{figure}

While the LFM ensures numerical stability through symmetric saturation, it applies identical attenuation to policy shifts of equal magnitude $|\log r|$ regardless of their direction. As established in the preceding analysis, the two shift directions carry fundamentally distinct risk profiles: overly aggressive probability amplification can induce mode collapse, whereas excessive probability erosion can precipitate catastrophic policy degradation. Symmetric treatment therefore fails to provide adequate direction-dependent regulation for these asymmetric failure modes. To address this limitation, we introduce the Decoupled Hazard Penalty (DHP), denoted as $\zeta(\cdot)$, which decouples the regulation of positive and negative policy shifts (see Figure.~\ref{fig:dhp_analysis} for an illustration). The transformation function $\zeta(\cdot)$ is defined as:
\begin{equation}
\zeta(\cdot) = \Big(\frac{s(\psi(\cdot))}{\lambda_{+}}\Big)^{k_+} + \Big(\frac{s(-\psi(\cdot))}{\lambda_{-}}\Big)^{k_-},
\end{equation}
where $s(\cdot)$ denotes the standard softplus function $s(x) = \log(1 + e^x)$. Here, $(\lambda_+, k_+)$ and $(\lambda_-, k_-)$ are independent hyperparameter pairs dedicated to governing positive and negative policy shifts, respectively. 

The proposed DHP enables fine-grained, asymmetric control over the optimization landscape through two core mechanisms:
\begin{itemize}
    \item \textbf{Directional Penalty Decoupling.} We leverage the softplus function to separate the penalty mechanisms for positive and negative shifts while maintaining differentiability. The decoupling is achieved by splitting the modulated log-ratio $\psi(\cdot)$ into two mirrored components. For positive shifts where $r_t^i(\theta) > 1$ and $\psi(r_t^i(\theta)) > 0$, the term $s(\psi(r_t^i(\theta)))$ approaches the identity function $\psi(r_t^i(\theta))$, while its counterpart $s(-\psi(r_t^i(\theta)))$ decays exponentially toward zero. Consequently, the penalty is predominantly governed by the term $(s(\psi(\cdot))/\lambda_{+})^{k_+}$. Conversely, for negative shifts ($r_t^i(\theta) < 1$), the term $s(-\psi(r_t^i(\theta)))$ becomes dominant, shifting the penalty control to $(s(-\psi(\cdot))/\lambda_{-})^{k_-}$. This mechanism ensures that the penalty is dominated by the term aligned with the shift direction, while the opposite term remains negligible, achieving a decoupled regulatory effect.
    \item \textbf{Hazard-aware Penalty Shaping.} To dynamically modulate the severity of these penalties, we adopt a hazard-aware shaping mechanism inspired by reliability theory and survival analysis. Specifically, we employ the cumulative hazard function of a Weibull distribution. For a non-negative input $x \ge 0$, this function is defined as:
    \begin{equation}
        H(x) = \left( \frac{x}{\lambda} \right)^k.
    \end{equation}
    In our proposed mechanism, the scale parameters $\lambda_{\pm}$ act as normalization constants. When the input $s(\psi(\cdot))$ reaches $\lambda_{\pm}$, the penalty achieves unit intensity, marking a critical threshold where the shape parameter $k_{\pm}$ begins to govern the growth rate. By setting $k_{\pm} > 1$, the penalty is effectively attenuated for $s(\psi(\cdot)) < \lambda_{\pm}$, decaying rapidly as $s(\psi(\cdot))$ moves toward zero. This characteristic ensures that minor deviations remain negligible, thereby facilitating safe exploration within a defined trust region. Conversely, for $s(\psi(\cdot)) > \lambda_{\pm}$, the penalty accelerates rapidly, strictly suppressing large policy deviations that could destabilize the system.
\end{itemize}

By assigning distinct parameter sets $(\lambda_+, k_+)$ and $(\lambda_-, k_-)$, the proposed DHP can effectively balance the trade-off between policy exploration and collapse, ensuring robust performance across diverse shift dynamics. Specifically, positive shifts can be lightly regulated to encourage the exploitation of high-reward trajectories, while negative shifts can be strictly penalized with a lower $\lambda_-$ or higher $k_-$ to prevent catastrophic policy collapses. The complete training procedure is summarized in Algorithm~\ref{alg:shape}.

\subsection{Stability Analysis of \methodname{}}
\label{sec:stability}

In this section, we formalize the optimization dynamics of \methodname{} and provide theoretical guarantees for its training stability. The complete training procedure is detailed in Algorithm~\ref{alg:shape}. 

To prevent the potential gradient reversal and instability that can arise when backpropagating through the DHP, we optimize the \methodname{} objective using a semi-gradient approach. Specifically, a stop-gradient operator is applied to the LFM output $\psi(r_t^i(\theta))$ within the penalty function $\zeta(\cdot)$. This design choice ensures that $\zeta(\cdot)$ acts strictly as a hazard-aware modulator, which rescales the update magnitude without distorting the direction of the policy gradient. 

Under this semi-gradient scheme, we analyze the stability of the \methodname{} objective defined in Eq.~\eqref{equation:methodname_loss}. The token-level objective is denoted as:
\begin{equation}
J_{\methodname{}}(r_t^i(\theta)) = \exp\big(\psi(r_t^i(\theta)) - \text{sg}[\zeta(r_t^i(\theta))]\big) \hat{A}^{i}_t,
\end{equation}
where $\text{sg}[\cdot]$ denotes the stop-gradient operator. 
For simplicity, we write $\zeta(\cdot)$ as shorthand for $\text{sg}[\zeta(\cdot)]$. Consequently, the gradient of $J_{\methodname{}}(r_t^i(\theta))$ with respect to $\theta$ is derived as:
\begin{align}
    \nabla_\theta J_{\methodname{}}(r_t^i(\theta)) & =  \exp(\psi(r_t^i(\theta)) - \zeta(r_t^i(\theta))) \cdot \nabla_\theta \psi(r_t^i(\theta)) \cdot \hat{A}^{i}_t \nonumber \\
    & = \exp(\psi(r_t^i(\theta)) - \zeta(r_t^i(\theta))) \cdot \frac{d\psi}{dr_t^i} \cdot \nabla_\theta r_t^i(\theta) \cdot \hat{A}^{i}_t \nonumber \\ 
    & = \exp(\psi(r_t^i(\theta)) - \zeta(r_t^i(\theta))) \cdot \underbrace{\left[\frac{1}{r_t^i(\theta)} \sech^2\left(\frac{\log r_t^i(\theta)}{c}\right)\right]}_{\text{LFM derivative}} \cdot r_t^i(\theta)\nabla_\theta \log \pi_\theta(q_t^i \mid p, q_{<t}^i) \cdot \hat{A}^{i}_t \nonumber \\
    & =  \underbrace{\exp(\psi(r_t^i(\theta)) - \zeta(r_t^i(\theta)))\sech^2\left(\frac{\log r_t^i(\theta)}{c}\right)}_{\mathcal{M}(r_t^i(\theta))}\nabla_\theta \log \pi_\theta(q_t^i \mid p, q_{<t}^i) \cdot \hat{A}^{i}_t, \label{equation:methodname_gradient}
\end{align}
where $\mathcal{M}(r)$ is defined as the gradient multiplier, which smoothly rescales the gradient at each token.

\begin{algorithm}[t]
    \caption{\methodname{} Training}
    \label{alg:shape}
    \begin{algorithmic}[1]
    \REQUIRE Policy $\pi_\theta$, dataset $\cD$, LFM bound $c$, DHP parameters $(k_+, \lambda_+, k_-, \lambda_-)$
    \FOR{each iteration}
        \STATE Sample prompts $p$ from $\cD$; generate $K$ responses $\{q^1, \ldots, q^K\}$ per prompt using $\pi_\theta$
        \STATE Compute group-normalized advantages $\hat{A}^{i}_t$
        \FOR{each response $q^i$, each token $t$}
            \STATE $r_t^i(\theta) \gets \pi_\theta(q_t^i \mid p, q_{<t}^i) \,/\, \pi_{\theta_{\text{old}}}(q_t^i \mid p, q_{<t}^i)$ \hfill \textit{// importance ratio}
            \STATE $\psi_t \gets c \cdot \tanh\!\big(\log r_t^i(\theta) \,/\, c\big)$ \hfill \textit{// LFM (Eq.~\eqref{equation:psi_zeta_definition})}
            \STATE $\zeta_t \gets \big(s(\sg{\psi_t}) / \lambda_+\big)^{k_+} + \big(s(-\sg{\psi_t}) / \lambda_-\big)^{k_-}$ \hfill \textit{// DHP (stop-grad on $\psi$)}
        \ENDFOR
        \STATE $\cL \gets -\frac{1}{K}\sum_{i}\frac{1}{T_i}\sum_{t} \exp(\psi_t - \zeta_t) \cdot \hat{A}^{i}_t$; update $\theta$ via gradient descent
    \ENDFOR
    \end{algorithmic}
\end{algorithm}

To evaluate the gradient stability of the \methodname{} objective, we first demonstrate that $\mathcal{M}(r_t^i(\theta))$ admits a finite upper bound, regardless of how extreme the policy shift $r_t^i(\theta)$ becomes. Given that $\zeta(\cdot) \ge 0$, the exponential penalty $\exp(-\zeta)$ acts as a contractive modulator ($\le 1$). Consequently, the supremum of the gradient multiplier can be bounded as follows:
\begin{equation}
    \sup_{r > 0}\; \big|\mathcal{M}(r)\big|
    \le
    \sup_{r > 0}\; \exp(\psi(r))\,\sech^2\!\left(\frac{\log r}{c}\right)
    \triangleq
    \hlbound{M_\psi(c)}
    =
    \frac{2}{1+\sqrt{1+c^2}}\,
    \exp\!\left(\frac{c^2}{1+\sqrt{1+c^2}}\right)
    \le
    e^{c}.
    \label{equation:M_bound}
\end{equation}

Here $\hlbound{M_\psi(c)}$ is the closed-form upper bound on the gradient multiplier induced by the given LFM mapping $\psi(\cdot)$. Building upon this bounded gradient multiplier, we analyze the statistical properties of the mini-batch gradient (Eq.~\eqref{equation:methodname_gradient}). In stochastic optimization, the second moment of the gradient estimator governs the reliability of the parameter updates and the stability of adaptive learning rate accumulators (e.g., Adam). By propagating the uniform bound from Eq.~\eqref{equation:M_bound} into the expectation of the mini-batch gradient, we establish the following theorem, which provides a formal upper bound on the mini-batch gradient estimator $g(\theta)$:
\begin{theorem}[Second-Moment Stability of the Mini-batch Gradient Estimator]
    \label{thm:second_moment_stability}
    Assume the group-relative advantage has variance $\E[(\hat{A}^{i}_t)^2]\le \sigma_A^2$, and the score function satisfies $\E[\|\nabla_\theta \log \pi_\theta(q_t^i \mid p, q_{<t}^i)\|^2]\le G^2$. The mini-batch gradient estimator $g(\theta)=\frac{1}{K}\sum_{i=1}^{K}\sum_{t=1}^{T_i}\mathcal{M}(r_t^i(\theta))\nabla_\theta \log \pi_\theta(q_t^i \mid p, q_{<t}^i)\hat{A}^{i}_t$ satisfies
    \begin{equation}
        \E\!\left[\|g(\theta)\|^2\right]
        \le
        \sigma_A^2\,G^2\,\hlbound{M_\psi(c)}^2
        \le
        \sigma_A^2\,G^2\,e^{2c}.
        \label{eq:grad_control_main}
    \end{equation}
\end{theorem}
Theorem \ref{thm:second_moment_stability} indicates that the variance of the gradient estimator is strictly controlled by the scale parameter $c$. By ensuring a well-behaved second moment, $\methodname{}$ prevents the preconditioner in adaptive optimizers (e.g., Adam) from being distorted by outlier gradients. This theoretical guarantee ensures that the gradient signal remains statistically stable, effectively mitigating the risks of learning rate vanishing or policy erosion that often plague large-scale RL training.

\section{Experiments}
\label{sec:experiments}

We conduct an extensive empirical evaluation to validate the efficacy of \methodname{} across a diverse spectrum of model architectures and reasoning tasks.
Section~\ref{sec:exp_setup} delineates the experimental methodology, including the underlying models, datasets, and baseline algorithms.
Section~\ref{sec:main_results} presents the principal findings, systematically structured along key dimensions of model capability: base vs.\ instruction-tuned paradigms, generic vs.\ domain-specific priors, and text-only vs.\ vision-language modalities.
Finally, Section~\ref{sec:ablation} provides rigorous ablation studies on hyperparameter sensitivity and dissects the optimization stability through an analysis of gradient norm dynamics, reward trajectories, and checkpoint robustness.

\subsection{Experimental Setup}
\label{sec:exp_setup}

To rigorously assess the generalizability of \methodname{}, our experimental design covers multiple architectural paradigms, encompassing base vs.\ instruction-tuned models, generic vs.\ domain-specific initializations, and text-only vs.\ multimodal capacities.

\textbf{Models and Datasets.} For text-only mathematical reasoning, we instantiate our framework on Qwen3-4B-Base~\citep{yang2025qwen3}, Qwen2.5-7B-Instruct, and the domain-specialized Qwen2.5-Math-7B-Instruct~\citep{yang2024qwen2}.
The policy is optimized on the DAPO-Math-17k-Processed dataset~\citep{dapo2025} using verifiable outcome-based rewards. Zero-shot generalization is evaluated on rigorous competition-level benchmarks: AIME 2024/2025~\citep{aime}, AMC 2023~\citep{amc}, HMMT25~\citep{hmmt}, and MATH-500~\citep{math_benchmark}.
For vision-language mathematical reasoning, we employ Qwen2.5-VL-7B-Instruct~\citep{bai2025qwen25vl}. Optimization is performed on the Geometry3K dataset~\citep{lu2021intergps}, with out-of-domain evaluation conducted on MathVision~\citep{wang2024mathvision}, MathVista~\citep{lu2024mathvista}, and MathVerse~\citep{zhang2024mathverse} to assess multimodal spatial logic and geometric comprehension.

\textbf{Baselines and Metrics.} We benchmark \methodname{} against a suite of state-of-the-art policy optimization algorithms: GRPO~\citep{deepseekmath}, GSPO~\citep{gspo2025}, DAPO~\citep{dapo2025}, and SAPO~\citep{sapo2025}.
For robust evaluation, we sample $n=32$ responses per prompt during inference and report the expected pass rate (Avg@32).

\textbf{Implementation Details.} 
Unless otherwise specified, we configure the smooth bound parameter to $c=1.5$. The hazard parameters are initialized asymmetrically: positive Weibull parameters $(k_+,\lambda_+)=(1.5,1.0)$ and negative parameters $(k_-,\lambda_-)=(2.0,0.8)$, reflecting a heightened sensitivity to premature suppression.
To maintain computational efficiency, the maximum response length is constrained to 2{,}048 tokens for Qwen2.5 variants and 4{,}096 tokens for Qwen3-4B. The remainder of the training pipeline follows standard practices.

\subsection{Main Results}
\label{sec:main_results}

\begin{table*}[t]
\centering
\scriptsize
\setlength{\tabcolsep}{2pt}
\renewcommand{\arraystretch}{1.15}
\renewcommand\tabularxcolumn[1]{m{#1}}
\caption{Main results on instruction-tuned backbones. We report Avg@32 (\%) for the best checkpoint. Subtable~(A) reports results on \textbf{Qwen2.5-7B-Instruct} (general-purpose), Subtable~(B) reports text-only math reasoning on \textbf{Qwen2.5-Math-7B-Instruct} (domain-specialized), and Subtable~(C) reports vision-language math reasoning on \textbf{Qwen2.5-VL-7B-Instruct}.}
\label{tab:main_results_1}
\begin{tabularx}{\textwidth}{>{\raggedright\arraybackslash}p{0.22\textwidth} M H C P A G}
\toprule
\multicolumn{7}{l}{\textbf{(A) Qwen2.5-7B-Instruct}} \\
\midrule
\textbf{Method}
& \textbf{MATH500}
& \textbf{HMMT25}
& \textbf{AMC23}
& \textbf{AIME25}
& \textbf{AIME24}
& \textbf{Avg.} \\
\midrule
\textbf{Baseline} & 70.0\gainzero & 5.6\gainzero & 52.4\gainzero & 3.5\gainzero & 16.4\gainzero & 29.6\gainzero \\
GRPO~\citep{deepseekmath} & 79.2\gainp{9.2} & 7.8\gainp{2.2} & 68.3\gainp{15.9} & 8.6\gainp{5.1} & 27.4\gainp{11.0} & 38.3\gainp{8.7} \\
GSPO~\citep{gspo2025} & 80.6\gainp{10.6} & 8.5\gainp{2.9} & 70.1\gainp{17.7} & 10.8\gainp{7.3} & 26.8\gainp{10.4} & 39.4\gainp{9.8} \\
DAPO~\citep{dapo2025} & 81.2\gainp{11.2} & 9.1\gainp{3.5} & 71.3\gainp{18.9} & 11.4\gainp{7.9} & 28.0\gainp{11.6} & 40.2\gainp{10.6} \\
SAPO~\citep{sapo2025} & 80.4\gainp{10.4} & 8.0\gainp{2.4} & 69.2\gainp{16.8} & 10.1\gainp{6.6} & 27.1\gainp{10.7} & 39.0\gainp{9.4} \\
\textbf{\methodname{}} (Ours) & \textbf{82.1}\gainp{12.1} & \textbf{13.8}\gainp{8.2} & \textbf{74.6}\gainp{22.2} & \textbf{16.7}\gainp{13.2} & \textbf{31.5}\gainp{15.1} & \textbf{43.7}\gainp{14.1} \\
\bottomrule
\end{tabularx}

\vspace{0.35em} 

\begin{tabularx}{\textwidth}{>{\raggedright\arraybackslash}p{0.22\textwidth} M H C P A G}
\toprule
\multicolumn{7}{l}{\textbf{(B) Qwen2.5-Math-7B-Instruct}} \\
\midrule
\textbf{Method}
& \textbf{MATH500}
& \textbf{HMMT25}
& \textbf{AMC23}
& \textbf{AIME25}
& \textbf{AIME24}
& \textbf{Avg.} \\
\midrule
\textbf{Baseline} & 73.1\gainzero & 7.5\gainzero & 56.3\gainzero & 7.7\gainzero & 15.1\gainzero & 31.9\gainzero \\
GRPO~\citep{deepseekmath} & 82.3\gainp{9.2} & 9.6\gainp{2.1} & 72.4\gainp{16.1} & 11.0\gainp{3.3} & 30.8\gainp{15.7} & 41.2\gainp{9.3} \\
GSPO~\citep{gspo2025} & 83.8\gainp{10.7} & 10.9\gainp{3.4} & 74.6\gainp{18.3} & 14.2\gainp{6.5} & 30.1\gainp{15.0} & 42.7\gainp{10.8} \\
DAPO~\citep{dapo2025} & 84.4\gainp{11.3} & 11.5\gainp{4.0} & 75.8\gainp{19.5} & 15.0\gainp{7.3} & 31.2\gainp{16.1} & 43.6\gainp{11.7} \\
SAPO~\citep{sapo2025} & 83.6\gainp{10.5} & 10.0\gainp{2.5} & 72.5\gainp{16.2} & 13.1\gainp{5.4} & 30.0\gainp{14.9} & 41.8\gainp{9.9} \\
\textbf{\methodname{}} (Ours) & \textbf{85.2}\gainp{12.1} & \textbf{17.9}\gainp{10.4} & \textbf{79.3}\gainp{23.0} & \textbf{21.6}\gainp{13.9} & \textbf{35.1}\gainp{20.0} & \textbf{47.8}\gainp{15.9} \\
\bottomrule
\end{tabularx}

\vspace{0.35em} 

\begin{tabularx}{\textwidth}{>{\raggedright\arraybackslash}p{0.22\textwidth} T V E G}
\toprule
\multicolumn{5}{l}{\textbf{(C) Qwen2.5-VL-7B-Instruct}} \\
\midrule
\textbf{Method}
& \textbf{MathVista}
& \textbf{MathVision}
& \textbf{MathVerse}
& \textbf{Avg.} \\
\midrule
\textbf{Baseline} & 67.8\gainzero & 25.2\gainzero & 49.5\gainzero & 47.5\gainzero \\
GRPO~\citep{deepseekmath} & 68.6\gainp{0.8} & 27.3\gainp{2.1} & 52.7\gainp{3.2} & 49.5\gainp{2.0} \\
GSPO~\citep{gspo2025} & 69.1\gainp{1.3} & 27.4\gainp{2.2} & 52.1\gainp{2.6} & 49.5\gainp{2.0} \\
DAPO~\citep{dapo2025} & 71.2\gainp{3.4} & 29.1\gainp{3.9} & 56.5\gainp{7.0} & 52.3\gainp{4.8} \\
SAPO~\citep{sapo2025} & 70.7\gainp{2.9} & 28.6\gainp{3.4} & 56.3\gainp{6.8} & 51.9\gainp{4.4} \\
\textbf{\methodname{}} (Ours) & \textbf{71.8}\gainp{4.0} & \textbf{30.5}\gainp{5.3} & \textbf{56.7}\gainp{7.2} & \textbf{53.0}\gainp{5.5} \\
\bottomrule
\end{tabularx}
\end{table*}

We present a comprehensive empirical validation of \methodname{} across diverse architectural backbones.
Table~\ref{tab:main_results_1} reports the performance of instruction-tuned models, covering (1) Qwen2.5-7B-Instruct, (2) Qwen2.5-Math-7B-Instruct, and (3) Qwen2.5-VL-7B-Instruct.
Complementarily, Table~\ref{tab:base_results} evaluates foundation models absent prior instruction tuning.
Empirically, \methodname{} consistently achieves state-of-the-art results across all configurations, confirming the broad applicability and robustness of our decoupled fidelity and damping objective.

\textbf{Domain-Specialized vs.\ General-Purpose Models.}
On the general-purpose Qwen2.5-7B-Instruct backbone (Table~\ref{tab:main_results_1}~A), \methodname{} yields a macro-average of 43.7\%, outperforming the strongest baseline, DAPO, by an absolute margin of 3.5 points.
Notably, this performance advantage is amplified when applied to domain-specialized architectures.
Table~\ref{tab:main_results_1}~B illustrates results on Qwen2.5-Math-7B-Instruct, which has been extensively pre-trained on mathematical corpora.
Here, \methodname{} achieves a macro-average of 47.8\%, exceeding DAPO by 4.2 points.
Crucially, these gains are most pronounced on highly challenging, competition-level benchmarks: on HMMT25, \methodname{} reaches 17.9\% compared to DAPO's 11.5\% ($+6.4$ points), and on AIME25, it attains 21.6\% versus 15.0\% ($+6.6$ points).
These findings indicate that our stable optimization framework confers substantial enhancements even when the underlying model already exhibits formidable domain-specific reasoning priors.

\textbf{Text-Only vs.\ Vision-Language Architectures.}
Table~\ref{tab:main_results_1}~C assesses the efficacy of \methodname{} on multimodal mathematical reasoning tasks using Qwen2.5-VL-7B-Instruct.
\methodname{} achieves a macro-average of 53.0\%, yielding a substantial improvement of 5.5 points over the zero-shot baseline and surpassing DAPO by 0.7 points.
The most significant gain is recorded on MathVision ($+5.3$ points over the baseline and $+1.4$ points over DAPO), a benchmark that inherently demands sophisticated geometric diagram comprehension and spatial logic.
This evidence verifies that our bounded gradient multiplier strategy seamlessly adapts to multimodal contexts, adeptly navigating the increased complexity and variance of the optimization landscape inherent to joint visual-textual reasoning.

\textbf{Base Models Without Instruction Tuning.}
Table~\ref{tab:base_results} reports the performance on Qwen3-4B-Base, a foundation model evaluated without the benefit of prior instruction tuning.
\methodname{} achieves a remarkable macro-average of 52.9\%, outperforming the leading baseline, GSPO, by 5.7 points.
The improvements on advanced competition benchmarks are particularly striking.
On AIME25, \methodname{} elevates the accuracy from 34.0\% (GSPO) to 41.4\% ($+7.4$ points), and similarly on AIME24, performance surges from 39.7\% to 44.7\% ($+5.0$ points).
These compelling results underscore the criticality of stable optimization for base models. During reinforcement learning, foundation models inevitably undergo more dramatic policy distribution shifts than their instruction-tuned counterparts; our framework successfully mitigates the ensuing gradient instability, effectively preventing the premature convergence that afflicts standard policy optimization variants.

\begin{table*}[t]
\centering
\scriptsize
\setlength{\tabcolsep}{2pt}
\renewcommand{\arraystretch}{1.15}
\caption{Results on base model without instruction tuning. We report Avg@32 (\%) for the best checkpoint on Qwen3-4B-Base. Avg. denotes the macro-average across benchmarks.}
\label{tab:base_results}
\renewcommand\tabularxcolumn[1]{m{#1}}
\begin{tabularx}{\textwidth}{>{\raggedright\arraybackslash}p{0.22\textwidth} M H C P A G}
\toprule
\textbf{Method}
& \textbf{MATH500}
& \textbf{HMMT25}
& \textbf{AMC23}
& \textbf{AIME25}
& \textbf{AIME24}
& \textbf{Avg.} \\
\midrule
\textbf{Baseline} & 57.2\gainzero & 3.8\gainzero & 41.3\gainzero & 5.7\gainzero & 8.8\gainzero & 23.4\gainzero \\
GRPO~\citep{deepseekmath} & 64.8\gainp{7.6} & 11.1\gainp{7.3} & 81.3\gainp{40.0} & 30.8\gainp{25.1} & 40.4\gainp{31.6} & 45.7\gainp{22.3} \\
GSPO~\citep{gspo2025} & 66.2\gainp{9.0} & 12.4\gainp{8.6} & 83.5\gainp{42.2} & 34.0\gainp{28.3} & 39.7\gainp{30.9} & 47.2\gainp{23.8} \\
DAPO~\citep{dapo2025} & 64.0\gainp{6.8} & 14.9\gainp{11.1} & 75.8\gainp{34.5} & 29.2\gainp{23.5} & 34.2\gainp{25.4} & 43.6\gainp{20.2} \\
SAPO~\citep{sapo2025} & 66.4\gainp{9.2} & 11.5\gainp{7.7} & 81.4\gainp{40.1} & 32.9\gainp{27.2} & 39.6\gainp{30.8} & 46.4\gainp{23.0} \\
\textbf{\methodname{}} (Ours) & \textbf{70.9}\gainp{13.7} & \textbf{19.4}\gainp{15.6} & \textbf{88.2}\gainp{46.9} & \textbf{41.4}\gainp{35.7} & \textbf{44.7}\gainp{35.9} & \textbf{52.9}\gainp{29.5} \\
\bottomrule
\end{tabularx}
\end{table*}

\subsection{Ablation Studies}
\label{sec:ablation}

We conduct rigorous ablation studies to isolate the individual contributions of key algorithmic components and hyperparameters within \methodname{}.
Our analysis explores hyperparameter sensitivity dynamics and provides a multifaceted examination of training stability.

\subsubsection{Hyperparameter Sensitivity}
\label{sec:ablation_hyper}

Table~\ref{tab:ablation} presents controlled ablation studies over the three primary hyperparameters of \methodname{}: the smooth bound $c$, the Weibull shape parameter $k$, and the Weibull scale parameter $\lambda$, which collectively operationalize the fidelity and damping principles introduced in Section~\ref{sec:objective}.
Specifically, $c$ governs \emph{fidelity} by determining the certified effective-ratio interval $[e^{-c},e^{c}]$, whereas $k$ and $\lambda$ jointly regulate \emph{damping} by controlling the severity and onset of tail attenuation applied to out-of-distribution log-ratio deviations.
All evaluations are conducted using Qwen2.5-VL-7B trained on Geo3K, with macro-average Avg@32 reported across MathVista, MathVision, and MathVerse.

For the sweeps over $k$ and $\lambda$, we adopt the full hyperparameter configuration from Table~\ref{tab:main_results_1}~C as the baseline and vary each parameter independently along a one-dimensional grid.
For the sweep over $c$, we employ a modified protocol in which the hazard terms are omitted, so that the contribution of the smooth bound can be evaluated in isolation from Weibull damping.
This design simultaneously identifies the optimal $c$ in the hazard-free setting and enables a direct comparison between the hazard-free variant and the full framework of Table~\ref{tab:main_results_1}~C, thereby quantifying the performance contribution of hazard shaping.

\begin{table*}[t]
\centering
\caption{Hyperparameter sensitivity of \methodname{} (Qwen2.5-VL-7B, Geo3K training).
We report macro-average Avg@32 (\%) across MathVista, MathVision, and MathVerse.
Bold denotes the best value among the reported settings.} %
\label{tab:ablation}
\small
\setlength{\tabcolsep}{5pt}
\renewcommand{\arraystretch}{1.15}
\begin{tabular}{@{}c r c r c r@{}}
\toprule
\multicolumn{2}{c}{\textit{(a) Bound $c$ (hazard-free)}}
& \multicolumn{2}{c}{\textit{(b) Shape $k$ \; ($c{=}1.5,\;\lambda{=}0.8$)}}
& \multicolumn{2}{c}{\textit{(c) Scale $\lambda$ \; ($c{=}1.5,\;k{=}3.0$)}} \\
\cmidrule(lr){1-2}\cmidrule(lr){3-4}\cmidrule(lr){5-6}
$c$ & Avg@32 & $k$ & Avg@32 & $\lambda$ & Avg@32 \\
\midrule
0.5 & 48.8 & 1.0 & 52.2 & 0.5 & 52.9 \\
1.0 & 47.5 & 1.5 & 51.4 & \textbf{0.8} & \textbf{53.0} \\
\textbf{1.5} & \textbf{49.7} & 2.5 & 52.3 & 1.0 & 52.2 \\
2.0 & 49.6 & \textbf{3.0} & \textbf{53.4} & 2.0 & 50.0 \\
\bottomrule
\end{tabular}
\end{table*}

The bound parameter $c$ controls the width of the unsaturated region, thereby determining the fidelity with which the original gradient signal is preserved.
As shown in Table~\ref{tab:ablation}(a), an overly restrictive bound such as $c{=}0.5$ narrows the ratio-operator range to $\omega(r)\in[0.61,1.65]$, leading to consistent underperformance due to excessive gradient saturation that reduces the informativeness of policy updates.
Performance improves as $c$ increases, reaching an optimum at $c{=}1.5$, where the ratio-operator range $\omega(r)\in[0.22,4.48]$ achieves a well-calibrated balance between gradient fidelity and update stability.
Beyond this point, performance plateaus for $c \ge 1.5$, indicating that once the ratio-operator support is sufficiently broad, explicit tail damping supplants saturation as the predominant mechanism for controlling deleterious gradient extremes.

The shape parameter $k$ governs the rate at which the damping penalty accelerates relative to the deviation magnitude.
When $k{=}1$, the hazard rate remains constant, whereas for $k{>}1$ the hazard rate increases monotonically with deviation magnitude, inducing an \emph{accelerating} damping effect that penalizes extreme deviations superlinearly.
Table~\ref{tab:ablation}(b) demonstrates that aggressive damping with $k \ge 2.5$ yields superior generalization, with $k{=}3.0$ achieving the highest average score.
This finding corroborates the core theoretical premise that effective optimization stabilization necessitates explicit tail attenuation beyond the mere bounding of ratio operators.
Although the smooth bounded operator at $k{=}1.0$ exhibits competitive performance, the substantial improvements observed for $k{>}1$ validate the necessity of superlinear tail penalization.

The scale parameter $\lambda$ determines the onset threshold at which damping becomes effective.
A smaller value of $\lambda$ triggers attenuation at lower log-ratio magnitudes, while a larger value delays the suppression of extreme policy deviations.
As reported in Table~\ref{tab:ablation}(c), $\lambda{=}0.8$ yields the best performance, with scores deteriorating progressively as $\lambda$ increases toward $2.0$.
This result confirms that early-onset damping is necessary to arrest harmful gradient excursions before they lead to systemic policy collapse.
An excessively low onset at $\lambda{=}0.5$ also slightly reduces performance relative to the optimum, as it suppresses moderately beneficial exploration and dilutes the effective learning signal.
Taken together, these findings substantiate the trade-off between fidelity and damping: the attenuation mechanism must activate promptly enough to forestall instability while remaining sufficiently permissive to preserve constructive policy updates.

\textbf{Relationship to default hyperparameters.}
The ablation above employs symmetric hazard parameters to isolate one-dimensional effects under a controlled setting.
The default configuration reported in Section~\ref{sec:exp_setup}, by contrast, adopts an asymmetric parameterization in which $k_+{=}1.5$, $\lambda_+{=}1.0$, $k_-{=}2.0$, and $\lambda_-{=}0.8$, separately regulating positive and negative policy shifts to account for their inherently different risk profiles.
This asymmetric design is selected to promote robust generalization across heterogeneous model-task combinations rather than to optimize performance on any single setting.
The default asymmetric configuration achieves 53.0\% on the VL benchmarks, as reported in Table~\ref{tab:main_results_1}~C, which is competitive with the best symmetric ablation result of 53.4\% in Table~\ref{tab:ablation}(b) at $k{=}3.0$.
This comparison confirms that the asymmetric design trades marginal single-task optimality for broader cross-task stability.

\begin{table*}[t]
\centering
\scriptsize
\setlength{\tabcolsep}{2pt}
\renewcommand{\arraystretch}{1.15}
\caption{Training stability comparison between Best and Latest checkpoints (Qwen3-4B-Base). We report Avg@32 (\%) with separate columns for the best checkpoint (Best) and the final checkpoint (Latest). $\Delta$ denotes the performance degradation from Best to Latest. \methodname{} exhibits minimal degradation, demonstrating superior training stability.}
\label{tab:best_latest}
\begin{tabular}{l cc cc cc cc cc cc c}
\toprule
\textbf{Method}
& \multicolumn{2}{c}{\textbf{MATH500}}
& \multicolumn{2}{c}{\textbf{HMMT25}}
& \multicolumn{2}{c}{\textbf{AMC23}}
& \multicolumn{2}{c}{\textbf{AIME25}}
& \multicolumn{2}{c}{\textbf{AIME24}}
& \multicolumn{2}{c}{\textbf{Avg.}}
& \textbf{$\Delta$} \\
\cmidrule(lr){2-3}\cmidrule(lr){4-5}\cmidrule(lr){6-7}\cmidrule(lr){8-9}\cmidrule(lr){10-11}\cmidrule(lr){12-13}
& Best & Latest & Best & Latest & Best & Latest & Best & Latest & Best & Latest & Best & Latest & \\
\midrule
GRPO & 64.8 & 58.9 & 11.1 & 8.8 & 81.3 & 72.5 & 30.8 & 23.3 & 40.4 & 24.2 & 45.7 & 37.5 & $-$8.2 \\
GSPO & 66.2 & 59.9 & 12.4 & 8.6 & 83.5 & 75.9 & 34.0 & 25.9 & 39.7 & 27.0 & 47.2 & 39.5 & $-$7.7 \\
DAPO & 64.0 & 47.5 & 14.9 & 7.0 & 75.8 & 66.3 & 29.2 & 17.9 & 34.2 & 18.4 & 43.6 & 31.4 & $-$12.2 \\
SAPO & 66.4 & 59.2 & 11.5 & 7.3 & 81.4 & 77.6 & 32.9 & 23.4 & 39.6 & 24.0 & 46.4 & 38.3 & $-$8.1 \\
\midrule
\textbf{\methodname{}} (Ours) & \textbf{70.9} & \textbf{70.1} & \textbf{19.4} & \textbf{19.4} & \textbf{88.2} & \textbf{88.2} & \textbf{41.4} & \textbf{40.8} & \textbf{44.7} & \textbf{41.8} & \textbf{52.9} & \textbf{52.1} & $-$\textbf{0.8} \\
\bottomrule
\end{tabular}
\end{table*}

\subsubsection{Training Stability Analysis}
\label{sec:ablation_stability}

We comprehensively evaluate training stability through three distinct yet interrelated analytical lenses: gradient norm dynamics, reward progression trajectories, and empirical checkpoint robustness.

\textbf{Gradient Norm Dynamics.}
The gradient norm trajectory illustrated in Figure~\ref{fig:overview}(b) provides direct empirical validation of our theoretical propositions regarding gradient multipliers.
Baseline algorithms exhibit frequent, high-variance gradient spikes throughout the optimization trajectory. SAPO, in particular, is afflicted by pronounced bursts that frequently exceed twice the baseline gradient magnitude.
In stark contrast, \methodname{} maintains strictly bounded and stable gradient norms throughout the entirety of training.
This empirically confirms that even when generation trajectories diverge substantially from the behavioral policy distribution, the bounded gradient multiplier effectively curtails extreme amplification. This observation strictly aligns with our theoretical non-explosion (second-moment) guarantees.

\textbf{Reward Progression.}
As depicted in Figure~\ref{fig:overview}(c), \methodname{} reliably accelerates convergence, securing higher expected rewards early in the training phase and strictly sustaining this advantage throughout the optimization process.
Conversely, baseline methodologies such as DAPO suffer from premature plateaus or exhibit severe reward degradation in later stages—a direct corollary of the unmitigated gradient instabilities identified above.
Coupled with the benchmark gains presented in Figure~\ref{fig:overview}(a), these dynamics establish a robust causal link between constrained gradient variance and superior policy generalization.

\textbf{Checkpoint Robustness.}
To rigorously quantify this stability advantage, Table~\ref{tab:best_latest} cross-examines the performance disparity between the optimal (peak) and terminal checkpoints on the Qwen3-4B-Base architecture.
Standard policy optimization baselines are highly susceptible to severe late-stage policy collapse: DAPO's macro-average precipitously drops from 43.6\% to 31.4\% (an absolute degradation of 12.2 points), while GRPO, GSPO, and SAPO manifest parallel degradations ranging from 7.7 to 8.2 points.
In stark contrast, \methodname{} exhibits a highly resilient degradation profile of merely 0.8 points (52.9\% $\to$ 52.1\%).
This remarkable performance retention unequivocally demonstrates that strictly enforcing a bounded gradient multiplier ($M < \infty$) successfully neutralizes the accumulation of destructive gradient variance that inherently catalyzes late-stage policy collapse. Consequently, \methodname{} dramatically reduces the reliance on exhaustive early-stopping heuristics, conferring significant practical reliability for large-scale RLHF pipelines.

\section{Conclusion}
\label{sec:conclusion}

We have presented \methodname{}, a principled framework that reformulates importance ratio control in reinforcement learning as a joint fidelity and damping design problem.
The central insight is that training stability is governed by the gradient multiplier, the scalar quantity that determines how the log-ratio rescales the score-function gradient.
\methodname{} achieves this through a smooth bounded operator that preserves gradient fidelity and a bounded ratio-operator range with controllable onset and acceleration for asymmetric treatment of positive and negative deviations.
Extensive experiments spanning general-purpose, domain-specialized, and vision-language models demonstrate that \methodname{} consistently achieves state-of-the-art performance on mathematical reasoning benchmarks with particularly pronounced gains.

\bibliography{youtu_bib}

\end{document}